\documentclass[10pt,twocolumn,letterpaper]{article}

\usepackage{iccv}
\usepackage{times}
\usepackage{epsfig}
\usepackage{graphicx}
\usepackage{amsmath}
\usepackage{amssymb}
\usepackage[utf8]{inputenc}
\usepackage[ruled,vlined]{algorithm2e}
\usepackage{caption}
\usepackage{subcaption}
\usepackage{cite}
\usepackage[export]{adjustbox}
\usepackage{authblk}
\usepackage[symbol*]{footmisc}

\newcommand{\chapternote}[1]{{%
  \let\thempfn\relax
  \footnotetext[0]{\emph{#1}}
}}
\include{pythonlisting}

\DeclareMathOperator*{\argmax}{argmax}
\DeclareMathOperator*{\argmin}{argmin}

\usepackage[breaklinks=true,bookmarks=false]{hyperref}

\iccvfinalcopy 
\def\iccvPaperID{****} 

\ificcvfinal\fi

\begin{document}

\title{ Interpreting Undesirable Pixels for Image Classification on Black-Box Models}

\author[]{Sin-Han Kang$^{1}$, Hong-Gyu Jung$^{1}$ and Seong-Whan Lee$^{2}$}
\affil[1]{Department of Brain and Cognitive Engineering, Korea University, Seoul, Korea}
\affil[2]{Department of Artificial Intelligence, Korea University, Seoul, Korea}
\affil[]{\textit {\{kangsinhan, hkjung00, sw.lee\}@korea.ac.kr}}
\maketitle
\thispagestyle{empty}

\begin{abstract}

   In an effort to interpret black-box models, researches for developing explanation methods have proceeded in recent years. Most studies have tried to identify input pixels that are crucial to the prediction of a classifier. While this approach is meaningful to analyse the characteristic of black-box models, it is also important to investigate pixels that interfere with the prediction. To tackle this issue, in this paper, we propose an explanation method that visualizes undesirable regions to classify an image as a target class. To be specific, we divide the concept of undesirable regions into two terms: (1) factors for a target class, which hinder that black-box models identify intrinsic characteristics of a target class and (2) factors for non-target classes that are important regions for an image to be classified as other classes. We visualize such undesirable regions on heatmaps to qualitatively validate the proposed method. Furthermore, we present an evaluation metric to provide quantitative results on ImageNet.
\end{abstract}
\chapternote {This work was supported by Institute for Information \& communications Technology Planning \& Evaluation(IITP) grant funded by the Korea government(MSIT) (No.2017-0-01779, A machine learning and statistical inference framework for explainable artificial intelligence)
} 
\section{Introduction}

The tremendous growth of deep networks has brought about the solvability of key problems in computer vision such as object classification \cite{Simonyan15},\cite{szegedy2015going} and object detection \cite{redmon2016you},\cite{girshick2014rich}. At the same time, the complexity of models has also increased, making it difficult for humans to understand the decisions of the model. To improve interpretability of black-box models, explanation methods have been proposed in terms of model inspection \cite{fong2017interpretable},\cite{sundararajan2017axiomatic}, \cite{nguyen2016synthesizing} and outcome explanation \cite{xu2015show}, \cite{montavon2017explaining}. These studies focus on visualizing crucial pixels for a model prediction. In other words, if we remove those pixels, the prediction accuracy is significantly decreased. 

\begin{figure}[t]
\begin{center}
	\begin{subfigure}[t]{0.91\linewidth}
   		\begin{subfigure}[t]{.11\linewidth}
			\includegraphics[width=0.77\linewidth]{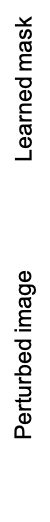}
		\end{subfigure}
		\begin{subfigure}[t]{.41\linewidth}
			\includegraphics[width=1\linewidth]{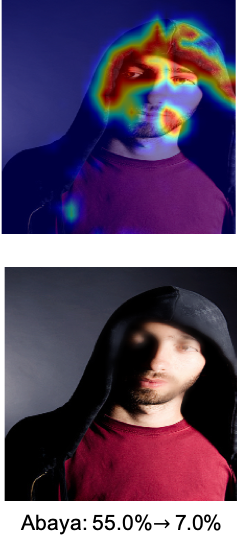}
    		\caption{ Local explanation\cite{fong2017interpretable}}
		\end{subfigure}
		\begin{subfigure}[t]{.422\linewidth}
 			\includegraphics[width=1\linewidth]{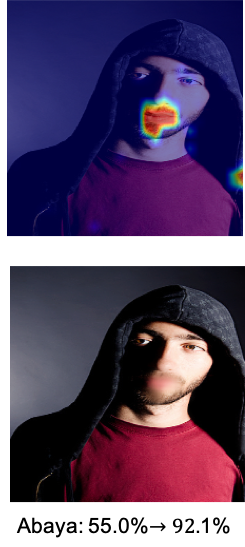}
 			\caption{Proposed method}
		\end{subfigure}
	\end{subfigure}
\caption{Comparison between a local explanation \cite{fong2017interpretable} and the proposed method. (a) \cite{fong2017interpretable} produces learned masks that are crucial to classify the abaya class. Thus, perturbing the pixels results in the accuracy significantly decreased. (b) the proposed method generates undesirable pixels for the abaya class. In this case, the perturbation image based on the learned mask improves the accuracy. The change on the accuracy before and after perturbation is presented on the bottom of each image.}
\label{fig:one}
\end{center}
\end{figure}

However, to obtain diverse interpretation on black-box models, it is also important to investigate pixels that interfere with the prediction. Thus, in this paper, we aim to find undesirable pixels that can improve the accuracy of a target class by perturbing the pixels. For example, Fig.~\ref{fig:one} shows the difference between \cite{fong2017interpretable} and our proposed method. While Fig.~\ref{fig:one}(a) explains that a hoodie and eyes play a major role to classify the image as an abaya that is a full-length outer garment worn by Muslim women, our method finds regions that help to improve the accuracy. Specifically, Fig.~\ref{fig:one}(b) interprets a mustache as undesirable pixels, which is generally not seen by women. Thus, perturbing the mustache leads to the accuracy improved for the abaya class. 

Figure ~\ref{fig:one} clearly shows that finding undesirable pixels of a target class can ease off the uncertainty about the decision of black-box models. Thus, we further define undesirable pixels with two different concepts. The first is factors for a target class (F-TC), which hinder that black-box models identify intrinsic characteristics of a target class. The second is factors for non-target classes (F-NTC) that are important regions for an image to be classified as other classes. In the following sections, we will mathematically elaborate on how these two different concepts interpret undesirable pixels for a model prediction. Then, we visually validate our idea on heatmaps and qualitatively evaluate the proposed method on ImageNet.

\section{Related Works}
Class activation map (CAM)\cite{zhou2014learning} and Grad-CAM \cite{selvaraju2017grad} analyze the decision of neural networks on heatmaps by utilizing activation maps of the last convolution layer in CNNs. Layer-wise relevance propagation (LRP) \cite{binder2016layer} computes gradients of the prediction score by exploiting a backward operation in neural networks. Model agnostic methods \cite{ribeiro2016should}, \cite{lundberg2017unified} approximate the perimeter decision boundary of a black-box model to a simple model such as logistic regression and decision tree. Local rule-based explanations (LORE) \cite{guidotti2018local} applys a genetic algorithm to build rule-based explanations, offering a set of counterfactual rules. Contrastive explanations methods (CEM)\cite{dhurandhar2018explanations} visualize a pertinent positive (PP) and a pertinent negative (PN) by using perturbation. But, PN is useful only when the meaning of classes for different inputs are similar to each other. Lastly, the most similar work to ours is local explanation\cite{fong2017interpretable} that learns a mask by perturbing important regions to a prediction. However, these methods do not clearly consider undesirable pixels of an image for a target class.

\section{Methods} \label{sect:methods}
Given an image $X\in\mathbb{R}^{H \times W \times 3}$, we generate a blurred image by applying Gaussian blur $h(X)=g_{\sigma,s}(X)$ where $\sigma$ and $s$ are standard deviation and kernel size, respectively. In order to replace specific pixels in $X$ with blurred pixels, we define a mask $M\in\left[ 0,1 \right]^{d}$, where $d$ is smaller than $H \times W$. Thus, a perturbed image is generated by the masking operator \cite{fong2017interpretable} as follows.
\begin{equation}
Q(X;M',h)=X \circ M'+h \left( X \right) \circ \left( 1-M' \right),
\end{equation}
where $ M' = Inp \left( M \right) $ is an interpolated mask and $Inp$ is a bilinear interpolation function. $\circ$ denotes the element-wise multiplication.
Given a black-box model ${f}$ and an accuracy $f_k\left(X\right)$ for a target class $k$, we expect that the perturbed image makes $f_k\left(X\right) \ll f_k\left( Q(X;M',h) \right)$. In other words, the goal is to find an optimal mask $M^{*}$ that improves the accuracy for a target class and an objective function can be defined as follows.
\begin{equation}
M^{*}=\argmax_M f_k\left( Q(X;M',h) \right).
\end{equation}
Since \cite{fong2017interpretable} shows that total-variation (TV) norm and $l1$ norm can produce reasonable and precise interpretability for the masking operation, we also apply such regularizers to our objective.
\begin{equation} \label{eq:4}
M^{*}=\argmax_M f_k\left( Q\left(X;M',h\right) \right) -\mathcal{R}_M, 
\end{equation}
where ${R}_M=\lambda_{1}\sum_{i,j}  ((M_{i+1,j}-M_{i,j})^{\beta} + 
(M_{i,j+1}-M_{i,j} )^{\beta} )^{{1 \over \beta}}+\lambda_{2}\lVert 1-M \rVert_{1}.$ $\lambda_{1}$, $\lambda_{2}$ and $\beta$
are hyper-parameters. 

\begin{figure}[t]
\begin{center}
   \includegraphics[width=1\linewidth]{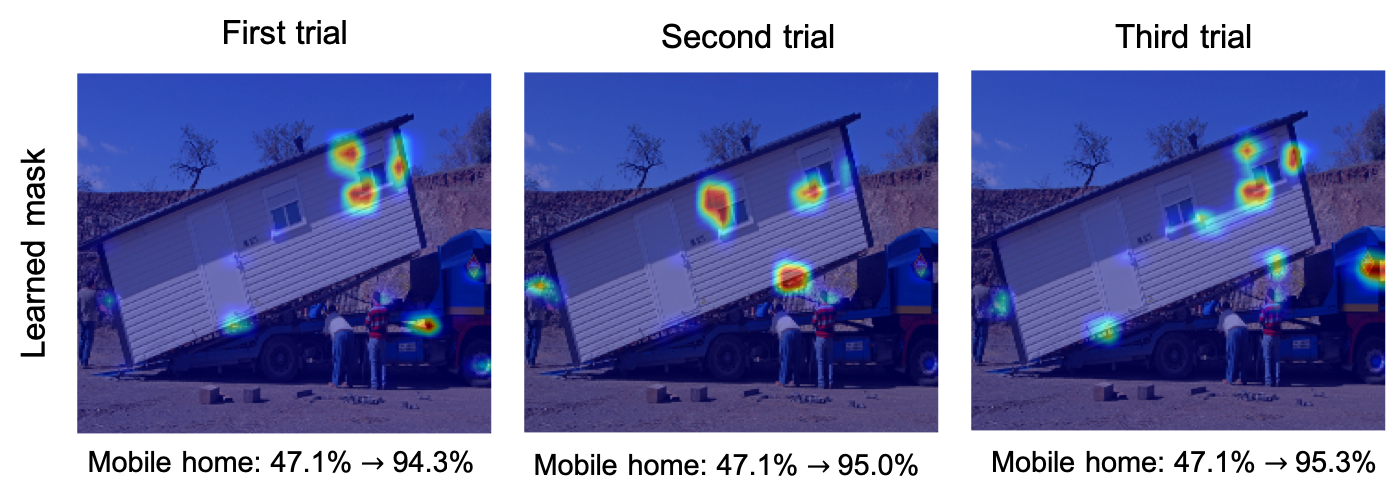}
   \caption{Learned masks generated by Eq.~\ref{eq:4} using TV and $l1$ norms. When applying Eq.~\ref{eq:4} several times, different learned masks are obtained. This leads to inconsistent interpretability.
   }
\label{fig:two}
\end{center}
\end{figure}

However, this objective function generates different masks for each trial as shown in Fig.~\ref{fig:two} and do not provide consistency of an explanation. We conjecture that this is due to the softmax operation. Given an output before softmax $y$, $f_k\left( Q \left(X;M',h \right)\right)={exp(y_k) \over \sum_{i} exp(y_i)} $ can be higher when increasing $exp(y_k)$ or decreasing $\sum_{i} exp(y_i)$. That is, improving $f_k\left( Q \left(X;M',h \right) \right)$ is affected by not only the output for a target class but also those for other classes. 

In order to solve this problem, we propose two types of regularizers for obtaining undesirable pixels. We first define factors for a target class (F-TC).
\begin{equation} \label{eq:F-TC}
\begin{split}
\mathcal{R}_{F-TC} = \gamma \lVert {1 \over N-1} \sum_{i,i\ne k} \{ & f_i' ( Q (X;M',h )) \\
				   & - f_i' ( X )\} \rVert_2,
\end{split}
\end{equation}
where $f_i' \left( \cdot \right)$ denotes the output before softmax for the $i$-th class, $k$ is the index of the target class, $\gamma$ is a hyper-parameter and $N$ is the total number of classes. This regularizer forces the objective function into focusing on the target class itself. In other words, $\mathcal{R}_{F-TC}$ finds the pixels that hinder intrinsic characteristic to be classified as the target class. The final objective function can be expressed as 
\begin{equation} 
\begin{split}
M^{*} = \argmax_M f_k \left( Q\left(X;M',h\right) \right) -\mathcal{R}_M 
- \mathcal{R}_{F-TC}.
\end{split}
\end{equation}
\begin{figure}[t]
\begin{center}
   \includegraphics[width=1\linewidth]{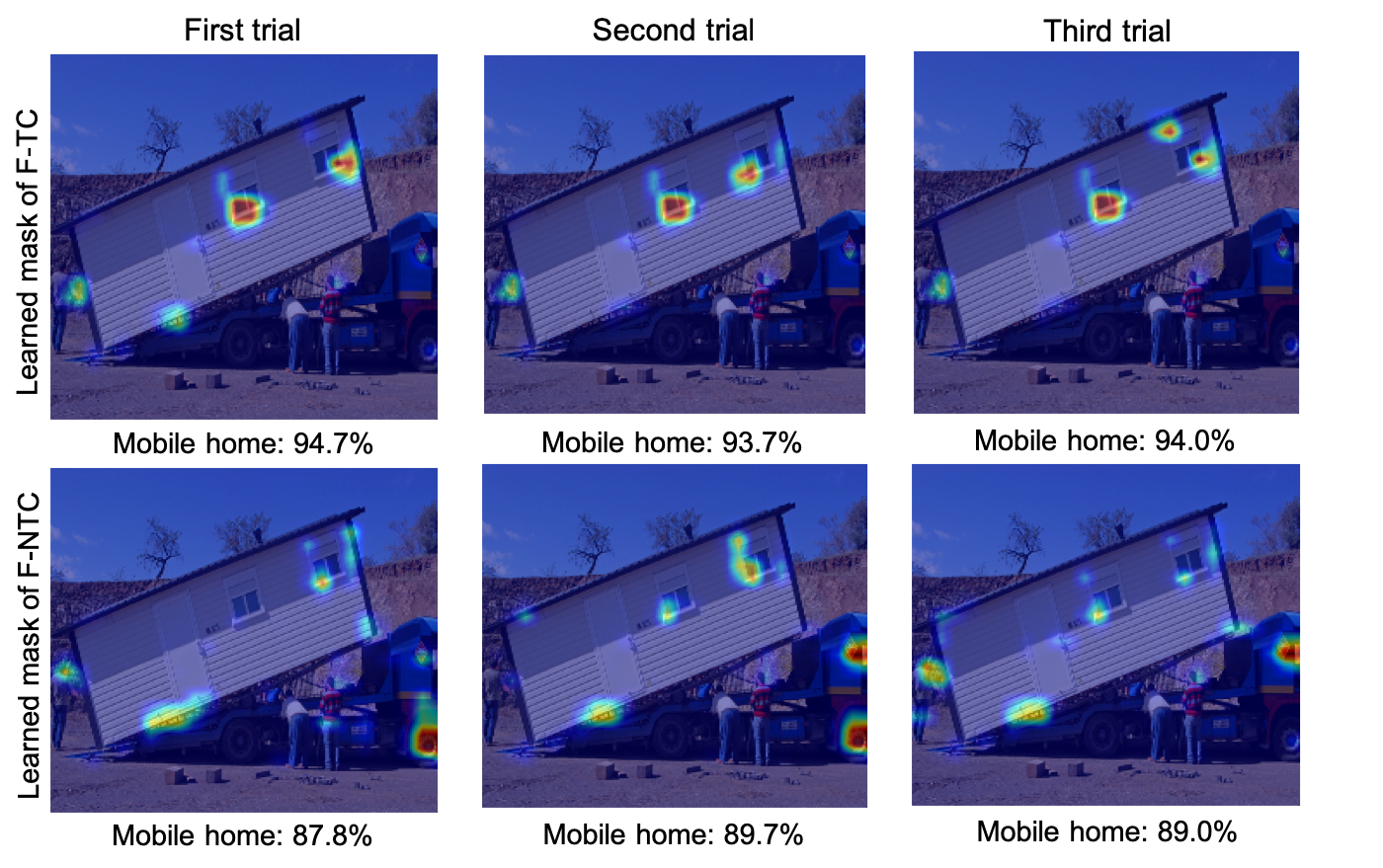}
   \caption{ Comparison of F-TC and F-NTC. F-TC explains that undesirable pixels to identify the mobile home are the windows. On the other hands, F-NTC focuses on the parts of the truck. Both methods produce consistent results for each trial. The accuracy of the target class is presented on the bottom of each image.
   }
   \label{fig:three}
\end{center}
\end{figure}
Secondly, we define factors for non-target classes (F-NTC).
\begin{equation} \label{eq:F-NTC}
\begin{split}
\mathcal{R}_{F-NTC} = \gamma\lVert \{ & f_k' ( Q (X;M',h ) ) \\
 & - f_{k}' (X) \} \rVert_2,
\end{split}
\end{equation}
which encourages to find undesirable pixels by focusing on other classes except for the target class. When applying $\mathcal{R}_{F-NTC}$, we modify Eq.~\ref{eq:4} as follows.
\begin{equation}
\begin{split}
M^{*} = \argmin_{M} \sum_{i, i\ne k} & \{ f_{i} (Q(X;M',h)) \}  \\ & +\mathcal{R}_{M} + \mathcal{R}_{F-NTC}.
\end{split}
\end{equation}

In the following section, we show several case studies to understand how these regularizations behave according to their definitions.

\section{Experiments}
\subsection{Experimental Settings}
We use VGG-19 \cite{Simonyan15} and ResNet-18 \cite{he2016deep} pretrained on the ImageNet \cite{russakovsky2015imagenet} and solve optimization problems using Adam \cite{kingma2014adam}. We set the learning rate to 0.1 and iterations to 200. We use the hyper-parameters $\lambda_1=1.7$, $\lambda_2 =3.0$, $\beta=2$ and $\gamma=0.3$. A mask 28$\times$28 is interpolated by 224 $\times$ 224 size by upsampling. The standard deviation $\sigma$ and kernel size $s$ for the Gaussian kernel are set to 5 and  11, respectively.
\begin{figure}[t]
\begin{center}
	\includegraphics[width=1\linewidth]{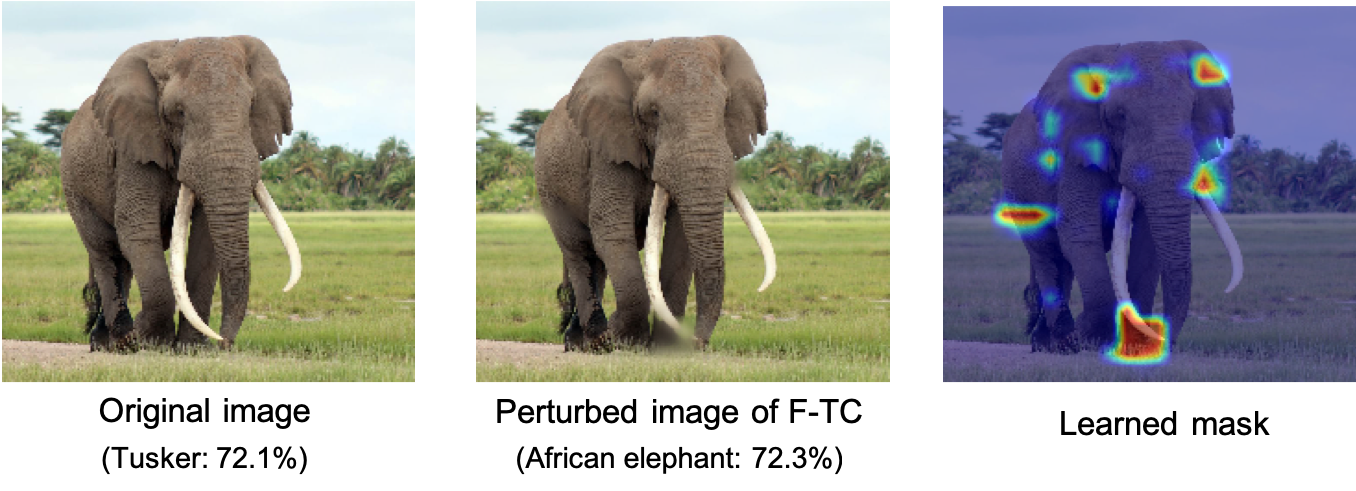}
   \caption{ 
   Visualizations of distinctive characteristics between similar classes. The length of the horn plays a major role to distinguish between an African elephant and a tusker. The highest accuracy for each class is presented by brackets.
   }
\label{fig:four}
\end{center}
\end{figure}
\subsection{Interpretability}
In Sect.~\ref{sect:methods}, we explained a main objective function with (1) TV norm and $l1$norms. Further, additional regularizers such as (2) F-TC and (3) F-NTC were proposed. We now compare interpretability among the three 
cases. First, as shown in Fig.~\ref{fig:two}, when merely using TV and $l1$ norms, the learned masks are generated irregularly for each trial. This makes us difficult to understand the decision of black-box models. On the other hands, Fig.~\ref{fig:three} shows that F-TC and F-NTC provide consistent visual interpretation. Moreover, each regularizer highlights the regions corresponding to their definitions such as Eq.~\ref{eq:F-TC} and Eq.~\ref{eq:F-NTC}. Specifically, F-TC explains that the windows are undesirable pixels to identify the intrinsic characteristic of the mobile home. F-NTC explains that the parts of the truck are undesirable pixels since those are more important to classify other classes such as a truck. In this way, our algorithm can be exploited to understand the decision of black-box models.
\\
\begin{figure*}
\begin{center}
\includegraphics[width=0.8\linewidth]{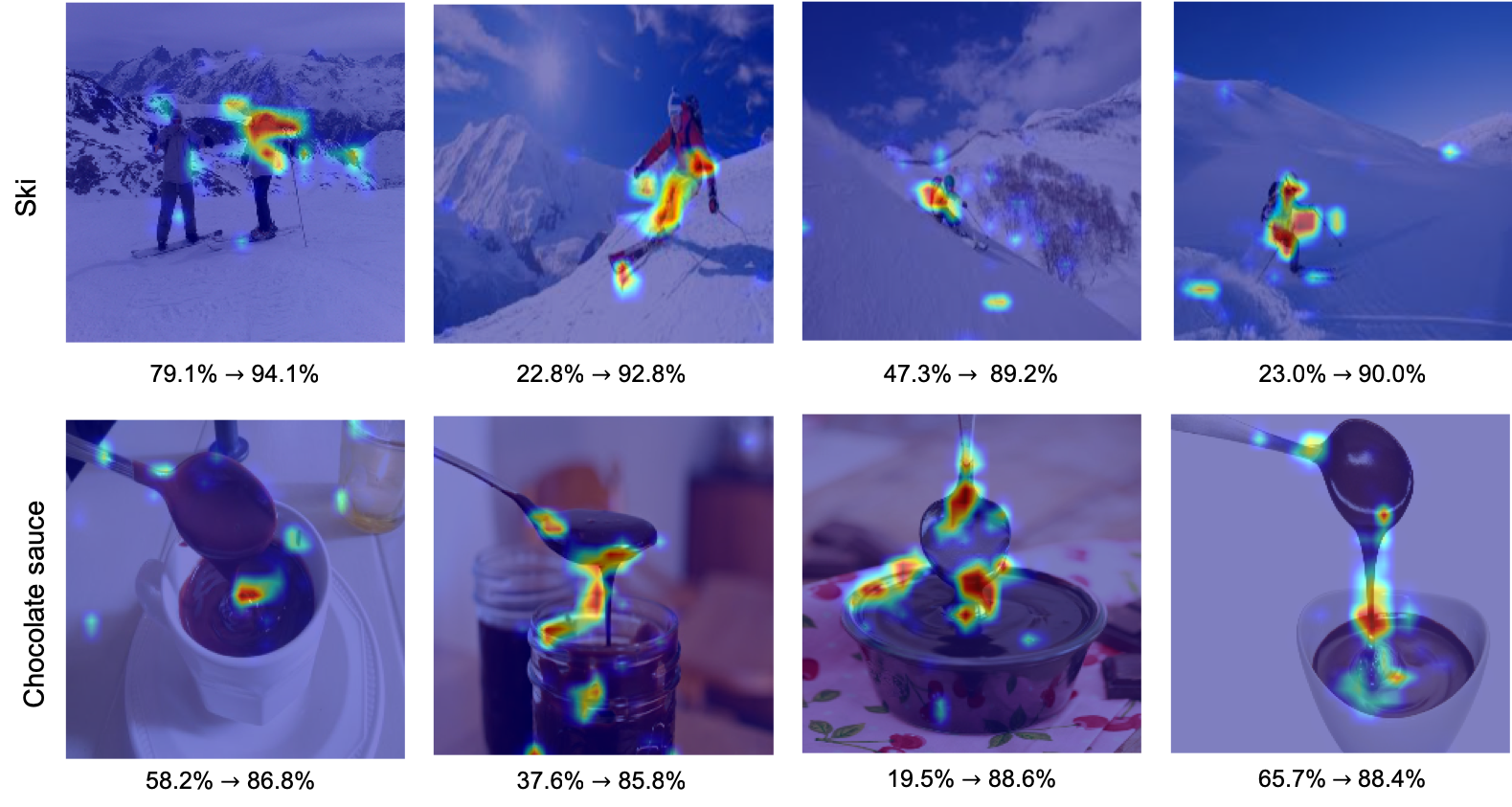}
   \caption{Behaviours of F-TC on VGG-19. The human body connected to ski equipments and the chocolate sauce falling into the cup are found as undesirable pixels to be classified as the target classes. The change on the accuracy before and after perturbation is presented on the bottom of each image.}
	\label{fig:five}
   \end{center}
\end{figure*}

\subsection{Qualitative Results}
We provide more examples to qualitatively evaluate the proposed method. We used VGG-19 for all experiments of this section.

As illustrated in Fig.~\ref{fig:four}, the original image is classified as the tusker with the accuracy of 72.1$\%$. When we set the African elephant as a target class, F-TC perturbs the end part of the horn, which results in improving the accuracy for the African elephant class. These results imply that the model generally regards the length of the horn as crucial features to distinguish between the tusker and the African elephant. More importantly, this is consistent with the fact that the horn of a tusker is longer than an African elephant. Thus, we argue that our method can provide reasonable interpretation about how trained networks distinguish similar classes.

Another example can be shown in Fig.~\ref{fig:five}. For ski and chocolate souce classes, F-TC highlights the human body connected to ski equipments and  the chocolate sauce that is falling from the spoon. These results suggest that portions connected to a target class have negative effect on a classification for a target class.

\subsection{Quantitative Results}
We present the following evaluation metric to measure how effectively our method finds undesirable pixels.
\begin{equation} \label{eq: evaluation}
\phi = \mathbb{E}_{X} \left[\left(f_h\left( Q\left(X;M',h \right) \right)-f_h(X)  \over 1-f_h(X) \right)*100\right],
\end{equation}
where $h$ is a class that has the highest accuracy for an image. $1-f_h(X)$ is the residual accuracy that can be improved from $f_h(X)$. Thus, Eq.~\ref{eq: evaluation} measures the relative accuracy improvement. We randomly select 1,000 images from the ImageNet and compare results between F-TC and F-NTC with VGG-19 and ResNet-18. In Table~\ref{tab:table1}, we observe that the accuracy can be effectively improved by perturbing undesirable pixels. We also measure the ratio of the number of undesirable pixels to the image size $224 \times 224$. In this case, we use the pixels that have magnitude above a threshold $0.6$. Table~\ref{tab:table2} shows that both F-TC and F-NTC yield a small number of undesirable pixels that are below 4$\%$. 
\begin{table}
\begin{center}
\begin{tabular}{|l||c|c|}
\hline
Model & F-TC & F-NTC \\
\hline\hline
VGG-19 & 48.741 & 48.979  \\
ResNet-18 & 44.898 & 44.688 \\
\hline
\end{tabular}
\caption{Relative accuracy improvement. The results indicate that perturbing undesirable pixels can effectively improve the classification performance.}
\label{tab:table1}
\end{center}
\end{table}

\begin{table}
\begin{center}
\begin{tabular}{|l||c|c|}
\hline
Model & F-TC & F-NTC \\
\hline\hline
VGG-19 & 0.0373 & 0.0378 \\
ResNet-18 & 0.0360 & 0.0398 \\
\hline
\end{tabular}
\caption{The percentage of undesirable pixels out of total image pixels. A small number of pixels are only used to find undesirable pixels.}
\label{tab:table2}
\end{center}
\end{table}

\section{Conclusion}
We proposed an explanation method that visualizes undesirable regions for classification. We defined the undesirable regions by two terms. The first is factors for a target class, which hinder that black-box models identify intrinsic characteristics of a target class and the second is factors for non-target classes that are important regions for an image to be classified as other classes. We showed the proposed method successfully found reasonable regions according to their definitions and by perturbing the undesirable pixels, we could improve the accuracy for the target class.
{\small
\bibliographystyle{ieee}
\bibliography{egbib}
}

\end{document}